# IMPROVEMENT OF BM3D ALGORITHM AND EMPLOYMENT TO SATELLITE AND CFA IMAGES DENOISING

Omid Pakdelazar[1] and Gholamali Rezai-rad [2]

[1]Department of Electrical and Electronic Engineering, Iran University of Science and Technology, Tehran, Iran.
`omidpakdelazar@gmail.com`
[2]Department of Electrical and Electronic Engineering, Iran University of Science and Technology, Tehran, Iran.
`rezai@iust.ac.ir`

*ABSTRACT*

*This paper proposes a new procedure in order to improve the performance of block matching and 3-D filtering (BM3D) image denoising algorithm. It is demonstrated that it is possible to achieve a better performance than that of BM3D algorithm in a variety of noise levels. This method changes BM3D algorithm parameter values according to noise level, removes prefiltering, which is used in high noise level; therefore Peak Signal-to-Noise Ratio (PSNR) and visual quality get improved, and BM3D complexities and processing time are reduced. This improved BM3D algorithm is extended and used to denoise satellite and color filter array (CFA) images. Output results show that the performance has upgraded in comparison with current methods of denoising satellite and CFA images. In this regard this algorithm is compared with Adaptive PCA algorithm, that has led to superior performance for denoising CFA images, on the subject of PSNR and visual quality. Also the processing time has decreased significantly.*

*KEYWORDS*

*Image denoising, improved BM3D, maximum d-distance, satellite image denoising, CFA image denoising.*

## 1. INTRODUCTION

Noise will be inevitably introduced in the image acquisition process; thus, denoising is an essential step to improve the image quality. Image denoising has been widely investigated as an initial image processing method during past four decades. Today, many schemes have been proposed to remove noise, from the earlier smoothing filters and frequency domain denoising methods [1], to bilateral filtering [2], Wavelet, and Multiresolution processing [3-8].

Recently, with the advancement of digital imaging equipments and due to wide range of applications, noise removal algorithms are needed to have high quality images.

In the transform –domain denoising procedure, it is assumed that noise –free signal can be well approximated by a linear compound of basic elements. Therefore, Image signal is represented in transform domain to be sparse. Hence, the noise–free image signal can be very good estimated by first keeping high-magnitude transform coefficients that have most of the signal energy and then removing the remaining coefficients that are caused by noise [9].

The multi-resolution transforms can represent localized details from image, like singularities and edges [9].

Multiresolution denoising algorithms are based on pointwise wavelet thresholding: its principle consists of setting all the wavelet coefficients to zero below a certain threshold value, while either keeping the remaining ones unchanged (hard-thresholding) or shrinking them by the threshold value (soft-thresholding, which is originally theorized by Donoho *et al.*) [7]. Some recently developed wavelet methods are curvele and ridgelet [10].

However, the overcompleteness transform alone cannot represent all of the image details. For instance, the discrete cosine transform (DCT) cannot be represented sharp edges and singularities, while wavelet transform is miserably accomplished in textures and smooth transitions. The great variety in natural images makes it impossible to achieve good sparsity for any fixed 2-D transform, for all cases. The adaptive principal components of local image patches is proposed by Muresan and Parks [9, 10] as a tool to overcome the mentioned drawbacks of standard orthogonal transforms. Lie Zahng *et al*. has extended and improved it on CFA image denoising [14]. However, when noise level is high, accurate estimation of the PCA basis is not possible and the image denoising performance is decreased. With similar intentions, the K-SVD algorithm [12] by Elad and Aharon utilizes highly overcomplete dictionaries obtained via a preliminary training procedure The weaknesses of PCA and learned dictionaries methods are the high computational burden they enforce to processors and the more time they need.

Recently, a powerful method for image denoisin by K. Dabov at al, based on block matching and 3-D Transform-Domain collaborative filtering (BM3D), is proposed [9].

This procedure proposed in transform domain improved sparse representation based a new image denoising method. The enhancement of the sparsity is achieved by grouping similar 2-D fragments of the image into 3-D data arrays, which we call "groups''. Collaborative filtering developed based on a special method for handling these 3- Dimensional groups. It includes three successive steps: 3-D transformation of a group, shrinkage of transform spectrum, and inverse 3-D transformation. Therefore, a 3- Dimensional group is obtained that comprise of a joint array from filtered 2-Dimensional fragments. Block Matching and 3-Dimensional filtering (BM3D) can achieve a high level of sparse representation of the noise –free signal, thus , the noise can be set apart well from signal by shrinkage. In this manner, the transform displays all of tiny details of image by grouped fractions, simultaneously the necessary unique feature of each individual fragment is protected.

Generally, denoising performance should gradually weaken with growing noise level. However, when noise standard deviation goes more than 39, denoising performance sharply drops. To avoid this problem, [9] proposes measuring the block-distance, using coarse prefiltering. It is shown in the following that by removing the prefiltering from the algorithm, its compatibility enhances [11]. Results show that by removal of prefiltering from BM3D algorithm and modification of parameters, such as maximum d-distance ($\tau_{match}^{ht}$), maximum number of grouped blocks ($N_2$), wiener filter parameter ($N_{step}$), PSNR and visual quality are augmented. The proposed method improves the output PSNR significantly even with the standard deviation less than 39.

This paper is divided into the following sections: sections 2 briefly surveys the concept of Block Matching and 3- Dimentional filtering (BM3D). In section 3 a method is proposed to improve the BM3D algorithm. Extension of the proposed algorithm for satellite images is presented in section 4 .Section 5 demonstrates the BM3D-based denoising algorithm for CFA images and the comparison of its performance with the PCA-based method. Finally, Section 6 is devoted to conclusions.

## 2. BLOCK MATCHING AND 3-D FILTERING

In this algorithm, the grouping is realized by block-matching, and the collaborative filtering is accomplished by shrinking in a 3-D transform domain. The used image fragments are square blocks of fixed size. The general procedure carried out in the algorithm is as follows: The input noisy image is processed by successive extraction of every reference block:

• Finding blocks that are similar to the reference one (block-matching), and stacking them together to form a 3-D array (group).

• Performing collaborative filtering of the group and returning the obtained 2-D estimates of all grouped blocks to their original locations.

After processing all reference blocks, the obtained block estimates can overlap; thus, there are multiple estimates for each pixel. These estimates are aggregated to shape a overall estimate of the entire image.

The general concept of the BM3D denoising algorithm is the following.

1. Block-wise estimates: For each block in the noisy image the filter performs:

(a) Grouping: Finding blocks that are similar to the currently processed one, and then stacking them together in a 3-D array (group).

The idea of the grouping is illustrated in Figure 1. Assuming that the stacked noisy blocks are corresponding to the perfectly identical noiseless blocks, an element-wise average (i.e. averaging between pixels at the same relative positions) will be an optimal estimator. The accuracy that can be achieved in this manner cannot be achieved with separate blocks independently. If all of the blocks in the same group are not exactly alike then averaging is not optimal anymore. Therefore, a filtering strategy, more effective than averaging, should be employed.

Figure 2 shows a demonstrative grouping sample by Block Matching and 3- Dimentional filtering, where a few reference blocks are shown and the ones matched as similar to them.

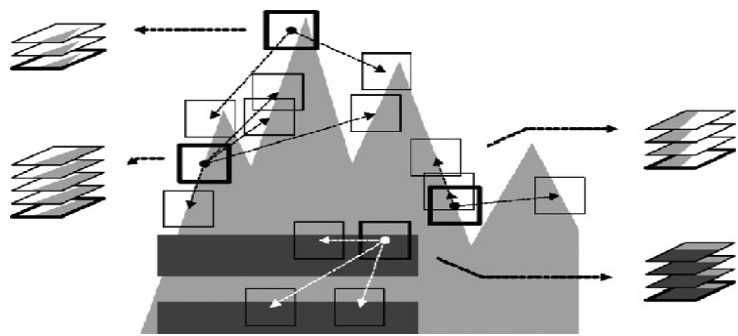

Figure 1. Simple example of grouping in an artificial image, where for each reference block (with thick borders) there exist perfectly similar ones [9].

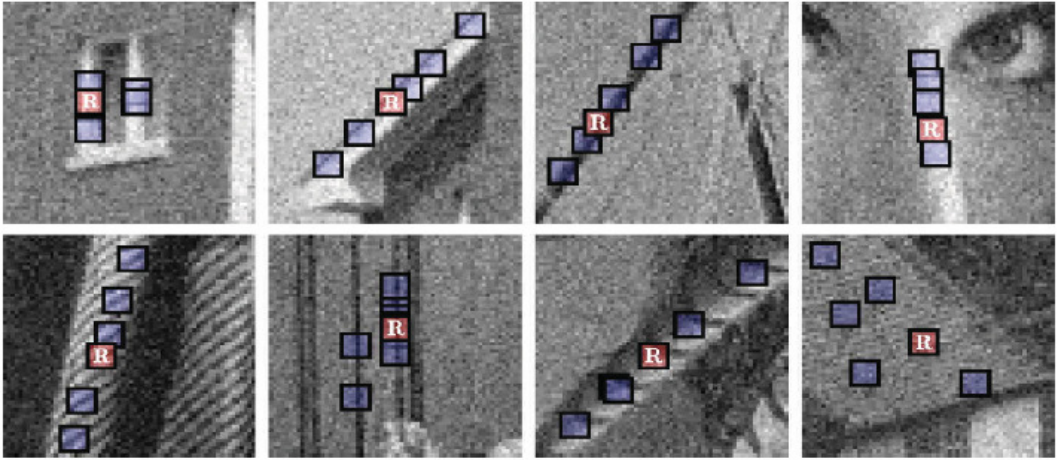

Figure. 2. Explanation of blocks that are grouped from noisy Image degraded by white Gaussian noise with zero mean and standard deviation 15.The reference blocks are marked with ''R'' and the rest of blocks are matched to it [9].

(b) Collaborative filtering:

Applying a 3-D transform to the formed group, attenuating the noise by shrinkage (e.g., hard-thresholding) of the transform coefficients, inverting the 3-D transform to produce estimates of all grouped blocks, and then returning the estimates of the blocks to their original places. Because the grouped blocks are similar, Block Matching and 3- Dimensional filtering (BM3D) can achieve a high level of sparse representation of the noise –free signal, thus, the noise can be set apart well from signal by shrinkage.

2. Aggregation. The output image is estimated by weighted averaging of all achieved block estimates that have overlap.

For a noisy image Z, one reference block ($Z_{x_R}$) within Z is determined. Grouping discovers the $Z_x$ that is similar to $Z_{x_R}$ by $\ell^2$-distance which can be calculated from the noisy blocks as

$$d^{noisy}(Z_{x_R}, Z_x) = \frac{\left\| Z_{x_R} - Z_x \right\|_2^2}{(N_1^{ht})^2} \quad (1)$$

Where $\|.\|$ denotes the $\ell^2$-norm and the blocks $Z_{x_R}$ and $Z_x$ are respectively located at $x_R$ and $x \in X$ in $Z$.

In [1], when noise standard deviation is more than 39 denoising performance has a sharp drop. To avoid this problem, it is proposed to measure the block-distance using a coarse prefiltering. This prefiltering is realized by applying a normalized 2-D linear transform on both blocks and then hard-thresholding the obtained coefficients, which results in

$$d(Z_{x_R}, Z_{x_R}) = \frac{\left\| \gamma'(\tau_{2D}^{ht}(Z_{x_R})) - \gamma'(\tau_{2D}^{ht}(Z_x)) \right\|_2^2}{(N_1^{ht})^2} \quad (2)$$

Where $\gamma'$ is the hard-thresholding operator with threshold $\lambda_{2D}\sigma$ and $\tau_{2D}^{ht}$ denotes the normalized 2-D linear transform.

Using the $\ell^2$-distance, the result of BM [1] is a set that contains the coordinates of the blocks that are similar to $Z_{x_R}$

$$S_{x_R}^{ht} = \{x \in X : d^{noisy}(Z_{x_R}, Z_x) \leq \tau_{match}^{ht}\} \quad (3)$$

Where the fixed $\tau_{match}^{ht}$ is the maximum $\ell^2$-distance for which two blocks are considered similar. These parameters are selected by deterministic conjectures (the acceptance value of the ideal distinction) it principally neglects the noisy components of the signal. Obviously $d^{noisy}(Z_{x_R}, Z_{x_R}) = 0$, which implies that $|S_{x_R}^{ht}| > 1$, where $|S_{x_R}^{ht}|$ denotes the cardinality of $S_{x_R}^{ht}$. After obtaining $S_{x_R}^{ht}$, a group is formed by stacking the matched noisy blocks $Z_{x \in S_{x_R}^{ht}}$ to form a 3-D array of size $N_1^{ht} \times N_1^{ht} \times |S_{x_R}^{ht}|$, which we denote $Z_{S_{x_R}^{ht}}$.

More details of BM3D algorithm are given in [9].

## 3. BM3D MODIFICATION

In [9], prefiltering is used to avoid sharp drop in denoising performance. However, using this filter will cause the removal of some noiseless data [11]. In the proposed method here, the algorithm parameters are set according to the added noise level to the image. Classifying the noise levels into low, medium, high and very high, the algorithm parameters can be set as mentioned in Table1.

As can be seen in Table 1, when the noise standard deviation is increased, the maximum size of grouped blocks $N_2$, should increase to improve the denoising performance. As mentioned in [9], $N_2$ is a power of 2 so it is allowed to change among 16, 32, 64 and so on.

Table 1. Comparison of parameter values of the improved BM3D and BM3D

| Standard deviation of noise added to the image | Changeable Parameter values of BM3D | Parameter value of improved BM3D |
|---|---|---|
| $\sigma < 30$ Low | $N_{step\_wiener} = 3$ <br> $\tau_{match}^{ht} = 2500$ | $N_{step\_wiener} = 2$ <br> $\tau_{match}^{ht} = 3000$ |
| $30 \leq \sigma < 50$ medium | $\tau_{match}^{ht} = 2500$ <br> $N_{step\_wiener} = 3$ | $\tau_{match}^{ht} = 6500$ <br> $N_{step\_wiener} = 2$ |
| $50 \leq \sigma < 80$ High | $N_1^{ht} = 11$ <br> $N_2 = 16$ <br> $\tau_{match}^{ht} = 5000$ | $N_1^{ht} = 8$ <br> $N_2 = 32$ <br> $\tau_{match}^{ht} = 15000$ |

---

[1] Block Matching

| $80 \leq \sigma \leq 100$ Very High | $N_1^{ht} = 11$ $N_2 = 16$ $\tau_{match}^{ht} = 5000$ | $N_1^{ht} = 8$ $N_2 = 64$ $\tau_{match}^{ht} = 30000$ |
|---|---|---|

$\tau_{match}^{ht}$ must be greater than its primitive value to ensure that there are enough blocks in 3D array for better image denoising performance. There is a trade-off between processing time and the output PSNR. Also, in noise level with standard deviation less than 50, wiener filter parameter $N_{step}$ should decrease from 3 to 2.

The output PSNR of the BM3D method [9], LPG-PCA method [10] and the proposed method are compared in Table 2 and in Figures (3), (4) and (5). PSNR is defined as follows:

$$PSNR = 10 \log_{10} (\frac{255^2}{MSE}) \qquad (4)$$

where MSE is the mean square error.

According to Table 2, it can be seen that the output PSNR values increase significantly with the noise level increment. Also, it can be seen in Figures (3), (4) and (5) that the proposed method shows more details than the two other methods.

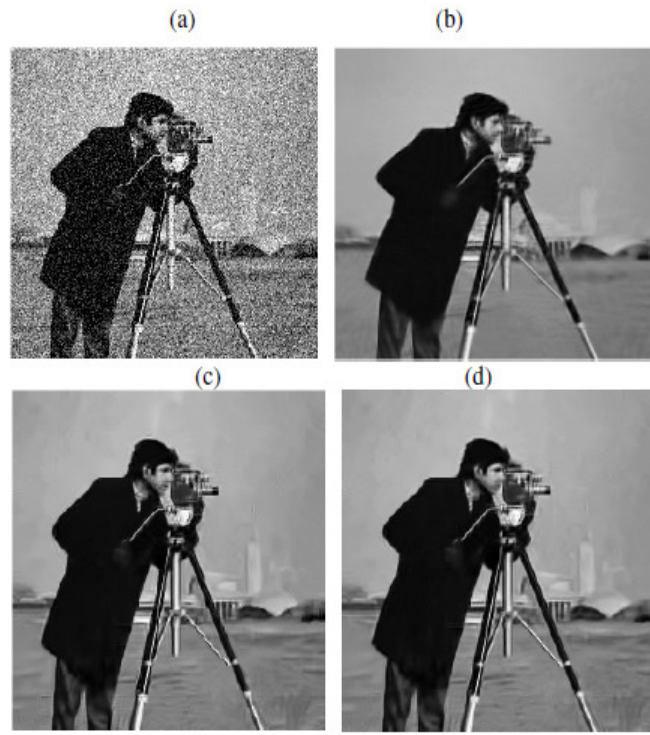

Figure 3. (a) Part of the noisy cameraman Image (PSNR=17.23 , $\sigma$ = 35). (b) Result of the LPG-PCA: PSNR=27.2. (c) Result of the BM3D: PSNR=27.83. (d) Result of the proposed method (improved BM3D): PSNR=28.01.

Table 2. The comparison of the output-PSNR of the proposed method with two state-of-the-art recent methods BM3D and LPG-PCA

| $\sigma$ / PSNR | house image 256*256 | | | Peppers image 256*256 | | |
|---|---|---|---|---|---|---|
| | LPG-PCA | BM3D | Improved BM3D | LPG-PCA | BM3D | Improved BM3D |
| 10 / 28.1 | 36.12 | 36.70 | **36.74** | 34.08 | 34.65 | **34.69** |
| 20 / 22.1 | 32.5 | 33.71 | **33.79** | 30.53 | 31.26 | **31.30** |
| 30 / 18.5 | 31.2 | 32.01 | **32.11** | 28.48 | 29.27 | **29.30** |
| 35 / 17.2 | 30.43 | 31.34 | **31.42** | 27.68 | 28.49 | **28.54** |
| 40 / 16.7 | 29.2 | 30.65 | **30.76** | 26.99 | 27.51 | **27.9** |
| 45 / 15.1 | 29.08 | 29.91 | **30.21** | 26.37 | 27.17 | **27.19** |
| 50 / 14.1 | 28.5 | 29.43 | **29.70** | 25.8 | 26.41 | **26.70** |
| 75 / 10.6 | 26.18 | 27.23 | **27.52** | 23.52 | 24.49 | **24.74** |
| 100 / 8.1 | 24.5 | 25.54 | **25.93** | 21.9 | 22.92 | **23.40** |

| $\sigma$ / PSNR | lena image 512*512 | | | boat image 512*512 | | |
|---|---|---|---|---|---|---|
| | LPG-PCA | BM3D | Improved BM3D | LPG-PCA | BM3D | Improved BM3D |
| 10 / 28.1 | 35.34 | 35.91 | **35.94** | 33.22 | 33.92 | **33.93** |
| 20 / 22.1 | 32.12 | 33.01 | **33.06** | 30.11 | 30.87 | **33.89** |
| 30 / 18.5 | 30.76 | 31.25 | **31.29** | 28.44 | 29.11 | **29.13** |
| 35 / 17.2 | 29.87 | 30.55 | **30.61** | 27.61 | 28.43 | **28.47** |
| 40 / 16.7 | 29.27 | 29.81 | **30.01** | 26.80 | 27.77 | **27.86** |
| 45 / 15.1 | 28.72 | 29.31 | **29.62** | 25.92 | 27.08 | **27.24** |
| 50 / 14.1 | 28.21 | 28.86 | **29.05** | 25.49 | 26.64 | **27.78** |
| 75 / 10.6 | 26.27 | 27.02 | **27.28** | 23.76 | 24.95 | **25.12** |
| 100 / 8.1 | 24.86 | 25.56 | **26.01** | 22.33 | 23.74 | **23.99** |

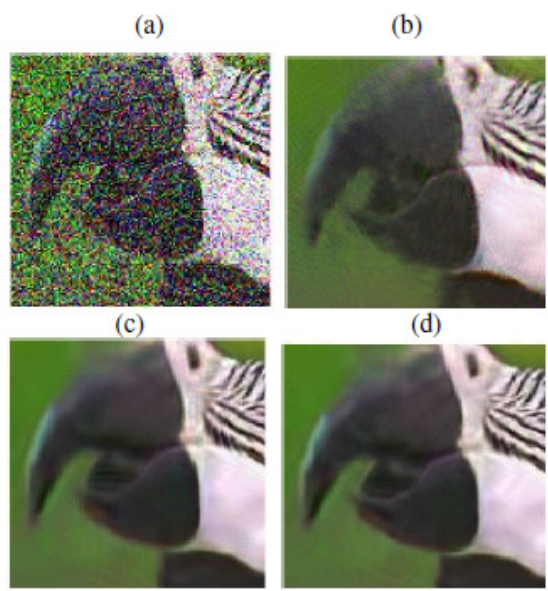

Figure 4. (a) Part of the noisy parrot Image (PSNR=10.64, $\sigma$ = 75). (b) Result of the LPG-PCA: PSNR=25.85. (c) Result of the BM3D: PSNR=27.25. (d) Result of the proposed method (improved BM3D): PSNR=27.56.

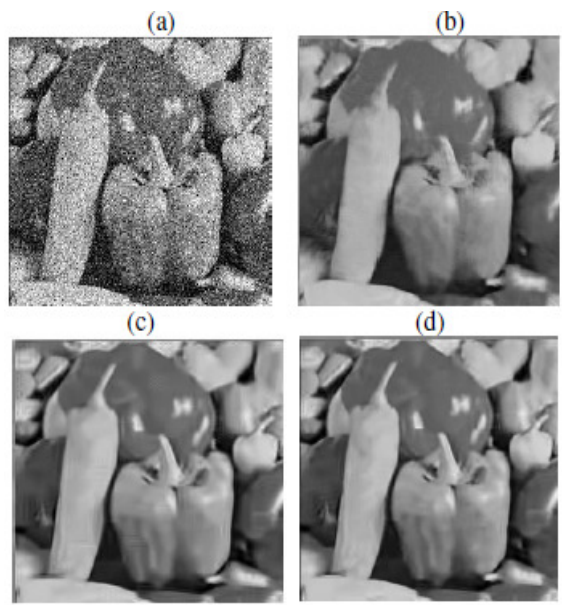

Figure 5. (a) Part of the noisy peppers Image (PSNR=14.13 , $\sigma$ =50). (b) Result of the LPG-PCA: PSNR=25.8. (c) Result of the BM3D: PSNR=26.41. (d) Result of the proposed method (improved BM3D): PSNR=26.70.

## 4. EXTENSION OF MODIFIED BM3D IN SATELLITE IMAGE DENOISING

Satellite image denoising requires much more time than standard images due to its voluminous data size and complexity. Improved BM3D can be applied to the satellite images in order to decrease the complexity and processing time. Because of the satellite images pixel size, threshold value $\tau_{match}^{ht}$ should increase by 3000 for all variety of noise levels. In addition to reduce processing time, N is changed from 11 to 8 and the Ns parameter (length of the side of the search neighborhood for full-search block-matching (BM)) is increased from 39 to 99.

The PSNR output of the proposed procedure is showed and is compared with two conventional methods. The implementation on the satellite images is illustrated in figure 6.

The results show that the output of the proposed algorithm on the satellite images has better performance than that of the state-of-the-art algorithms in terms of both Peak signal-to-noise ratio and visual quality.

## 5. APPLICATION OF MODIFIED BM3D IN CFA IMAGE DENOISING FOR SINGLE SENSOR DIGITAL CAMERAS

The output of the Single sensor digital cameras use a process called color demosaicking to produce full color images from data captured by a color filter array. Images quality degrades during image acquisition procedure as a result of sensor noise. Conventional solution for noise reduction is demosaicking first and then denoising the image. Color artifacts in the demosaicking process cause these conventional methods to produce many noise in image which will then make the noise removal process hard and difficult.

Recently a powerful method based on PCA is proposed for CFA image denoising. In this method, despite conventional methods, first denoising Is done with PCA and then the image is demosaicked. As mentioned before, the PCA algorithm consumes lots of time to implement and imposes a very high computational burden, therefore it has practical applications difficulties for digital cameras. To solve this problem, modified BM3D method can be used instead of PCA in denoising procedure. For simulation of CFA image with channel-dependent sensor noise, gaussian white noise is added separately with standard deviations of $\sigma_r = 30, \sigma_g = 27, \sigma_b = 25$ to red, green, and blue channels of the image, respectively. Then digital image resolution is decreased (downsampling). The PSNR of the output of the proposed method is compared with adaptive PCA method in Figure (7). It can be seen that the proposed method has better performance than the PCA method considering both PSNR and visual quality. In addition, computational processing time of the proposed algorithm is decreased significantly. The processing time for PCA and the proposed algorithms last 236sec and 39sec, respectively, on a X86-based computer with 2.8 GHz CPU on same conditions.

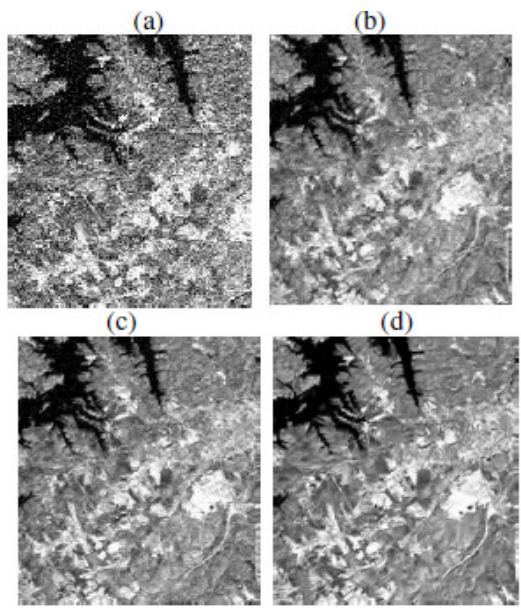

figure 6. (a) Parts of a noisy band of the Landsat image from [13]: (PSNR=17.23, $\sigma = 35$). (b) Result of [13]: PSNR=21.38. (c) Result of the [7]: PSNR=22.16. (d) Result of the proposed method: PSNR=22.41.

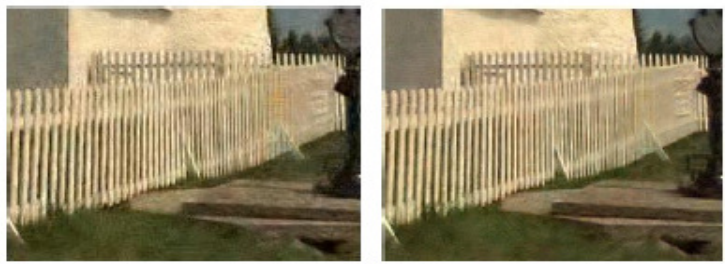

Figure 7. On the left: is reconstructed by the adaptive PCA-based CFA denoising method followed by demosaicking method [14]: PSNR= 27.61 dB. On the right: is reconstructed by the proposed BM3D-based CFA denoising method followed by demosaicking method: PSNR= 28.32 dB. (Standard deviation of noisy CFA data is ($\sigma_r = 30$, $\sigma_g = 27$, $\sigma_b = 25$)).

## 6. CONCLUSION

In this paper a modification of BM3D algorithm for image denoising was presented. The results show that by removal of prefiltering from BM3D algorithm and modifying parameters such as maximum d-distance ($\tau_{match}^{ht}$), maximum number of grouped blocks ($N_2$), wiener filter parameter ($N_{step}$), the PSNR and visual quality gets better than that of BM3D. This algorithm is extended to denoise satellite images, output results demonstrate that the output of the proposed algorithm on the satellite image has enhanced performance compared with the state-of-the-art algorithms in terms of both Peak signal-to-noise ratio and visual quality. Also when this algorithm is developed to CFA images denoising, it has superior PSNR and visual quality than adaptive PCA algorithms. In addition, it significantly reduces the processing time. Using the proposed method here in medical image processing can be considered as future research.

**Authors**

**Omid Pakdelazar** was born in Iran, in 1985.In 2009, he received the B.Sc. degree in electrical and electronics engineering from Garmsar University. He is currently pursuing the M.Sc. degree in electronics at Iran University of Science and Technology, Tehran, Iran. His research interests mainly include multiresolution analysis and the restoration of images.

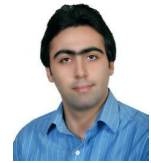

**Gholamali Rezai-rad** was born in Iran, in 1950. he received his B.Sc. degree in electrical and electronics engineering from University of District Columbia, USA. He received his M.Sc. and Ph.D. from George Washington University and is currently associated professor at Iran University of Science and Technology, Tehran, Iran. His research interests mainly include image processing, biomedical engineering and signals & systems.

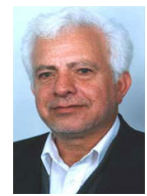